# Auto-Sizing Neural Networks:
# With Applications to $n$-gram Language Models

**Kenton Murray** and **David Chiang**
Department of Computer Science and Engineering
University of Notre Dame
`{kmurray4,dchiang}@nd.edu`

## Abstract

Neural networks have been shown to improve performance across a range of natural-language tasks. However, designing and training them can be complicated. Frequently, researchers resort to repeated experimentation to pick optimal settings. In this paper, we address the issue of choosing the correct number of units in hidden layers. We introduce a method for automatically adjusting network size by pruning out hidden units through $\ell_{\infty,1}$ and $\ell_{2,1}$ regularization. We apply this method to language modeling and demonstrate its ability to correctly choose the number of hidden units while maintaining perplexity. We also include these models in a machine translation decoder and show that these smaller neural models maintain the significant improvements of their unpruned versions.

## 1 Introduction

Neural networks have proven to be highly effective at many tasks in natural language. For example, neural language models and joint language/translation models improve machine translation quality significantly (Vaswani et al., 2013; Devlin et al., 2014). However, neural networks can be complicated to design and train well. Many decisions need to be made, and performance can be highly dependent on making them correctly. Yet the optimal settings are non-obvious and can be laborious to find, often requiring an extensive grid search involving numerous experiments.

In this paper, we focus on the choice of the sizes of hidden layers. We introduce a method for automatically pruning out hidden layer units, by adding a sparsity-inducing regularizer that encourages units to deactivate if not needed, so that they can be removed from the network. Thus, after training with more units than necessary, a network is produced that has hidden layers correctly sized, saving both time and memory when actually putting the network to use.

Using a neural $n$-gram language model (Bengio et al., 2003), we are able to show that our novel auto-sizing method is able to learn models that are smaller than models trained without the method, while maintaining nearly the same perplexity. The method has only a single hyperparameter to adjust (as opposed to adjusting the sizes of each of the hidden layers), and we find that the same setting works consistently well across different training data sizes, vocabulary sizes, and $n$-gram sizes. In addition, we show that incorporating these models into a machine translation decoder still results in large BLEU point improvements. The result is that fewer experiments are needed to obtain models that perform well and are correctly sized.

## 2 Background

Language models are often used in natural language processing tasks involving generation of text. For instance, in machine translation, the language model helps to output fluent translations, and in speech recognition, the language model helps to disambiguate among possible utterances.

Current language models are usually $n$-gram models, which look at the previous $(n-1)$ words to predict the $n$th word in a sequence, based on (smoothed) counts of $n$-grams collected from training data. These models are simple but very effective in improving the performance of natural language systems.

However, $n$-gram models suffer from some limitations, such as data sparsity and memory usage. As an alternative, researchers have begun exploring the use of neural networks for language modeling. For modeling $n$-grams, the most common approach is the feedforward network of Bengio et al. (2003), shown in Figure 1.

Each node represents a *unit* or "neuron," which has a real valued *activation*. The units are organized into real-vector valued *layers*. The activations at each layer are computed as follows. (We assume $n = 3$; the generalization is easy.) The two preceding words, $w_1, w_2$, are mapped into lower-dimensional word embeddings,

$$\mathbf{x}^1 = \mathbf{A}_{:w_1}$$
$$\mathbf{x}^2 = \mathbf{A}_{:w_2}$$

then passed through two hidden layers,

$$\mathbf{y} = f(\mathbf{B}^1 \mathbf{x}^1 + \mathbf{B}^2 \mathbf{x}^2 + \mathbf{b})$$
$$\mathbf{z} = f(\mathbf{C}\mathbf{y} + \mathbf{c})$$

where $f$ is an elementwise nonlinear *activation* (or *transfer*) function. Commonly used activation functions are the hyperbolic tangent, logistic function, and rectified linear units, to name a few. Finally, the result is mapped via a softmax to an output probability distribution,

$$P(w_n \mid w_1 \cdots w_{n-1}) \propto \exp([\mathbf{Dz} + \mathbf{d}]_{w_n}).$$

The parameters of the model are $\mathbf{A}$, $\mathbf{B}^1$, $\mathbf{B}^2$, $\mathbf{b}$, $\mathbf{C}$, $\mathbf{c}$, $\mathbf{D}$, and $\mathbf{d}$, which are learned by minimizing the negative log-likelihood of the the training data using stochastic gradient descent (also known as *backpropagation*) or variants.

Vaswani et al. (2013) showed that this model, with some improvements, can be used effectively during decoding in machine translation. In this paper, we use and extend their implementation.

## 3 Methods

Our method is focused on the challenge of choosing the number of units in the hidden layers of a feed-forward neural network. The networks used for different tasks require different numbers of units, and the layers in a single network also require different numbers of units. Choosing too few units can impair the performance of the network, and choosing too many units can lead to overfitting. It can also slow down computations with the network, which can be a major concern for many applications such as integrating neural language models into a machine translation decoder.

Our method starts out with a large number of units in each layer and then jointly trains the network while pruning out individual units when possible. The goal is to end up with a trained network

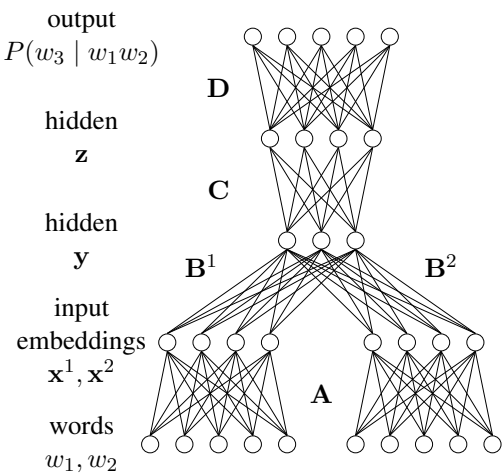

Figure 1: Neural probabilistic language model (Bengio et al., 2003), adapted from Vaswani et al. (2013).

that also has the optimal number of units in each layer.

We do this by adding a regularizer to the objective function. For simplicity, consider a single layer without bias, $\mathbf{y} = f(\mathbf{Wx})$. Let $L(\mathbf{W})$ be the negative log-likelihood of the model. Instead of minimizing $L(\mathbf{W})$ alone, we want to minimize $L(\mathbf{W}) + \lambda R(\mathbf{W})$, where $R(\mathbf{W})$ is a convex regularizer. The $\ell_1$ norm, $R(\mathbf{W}) = \|\mathbf{W}\|_1 = \sum_{i,j} |W_{ij}|$, is a common choice for pushing parameters to zero, which can be useful for preventing overfitting and reducing model size. However, we are interested not only in reducing the number of parameters but the number of units. To do this, we need a different regularizer.

We assume activation functions that satisfy $f(0) = 0$, such as the hyperbolic tangent or rectified linear unit ($f(x) = \max\{0, x\}$). Then, if we push the incoming weights of a unit $y_i$ to zero, that is, $W_{ij} = 0$ for all $j$ (as well as the bias, if any: $b_i = 0$), then $y_i = f(0) = 0$ is independent of the previous layers and contributes nothing to subsequent layers. So the unit can be removed without affecting the network at all. Therefore, we need a regularizer that pushes all the incoming connection weights to a unit together towards zero.

Here, we experiment with two, the $\ell_{2,1}$ norm and the $\ell_{\infty,1}$ norm.[1] The $\ell_{2,1}$ norm on a ma-

---

[1] In the notation $\ell_{p,q}$, the subscript $p$ corresponds to the norm over each group of parameters, and $q$ corresponds to the norm over the group norms. Contrary to more common usage, in this paper, the groups are rows, not columns.

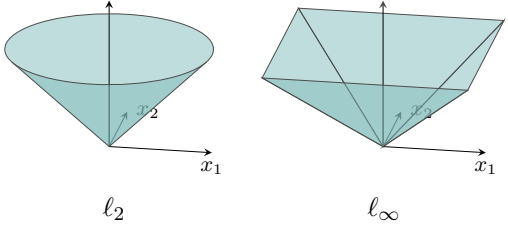

Figure 2: The (unsquared) $\ell_2$ norm and $\ell_\infty$ norm both have sharp tips at the origin that encourage sparsity.

trix $\mathbf{W}$ is

$$R(\mathbf{W}) = \sum_i \|W_{i:}\|_2 = \sum_i \left(\sum_j W_{ij}^2\right)^{\frac{1}{2}}. \quad (1)$$

(If there are biases $b_i$, they should be included as well.) This puts equal pressure on each row, but within each row, the larger values contribute more, and therefore there is more pressure on larger values towards zero. The $\ell_{\infty,1}$ norm is

$$R(\mathbf{W}) = \sum_i \|W_{i:}\|_\infty = \sum_i \max_j |W_{ij}|. \quad (2)$$

Again, this puts equal pressure on each row, but within each row, only the maximum value (or values) matter, and therefore the pressure towards zero is entirely on the maximum value(s).

Figure 2 visualizes the sparsity-inducing behavior of the two regularizers on a single row. Both have a sharp tip at the origin that encourages all the parameters in a row to become exactly zero.

## 4 Optimization

However, this also means that sparsity-inducing regularizers are not differentiable at zero, making gradient-based optimization methods trickier to apply. The methods we use are discussed in detail elsewhere (Duchi et al., 2008; Duchi and Singer, 2009); in this section, we include a short description of these methods for completeness.

### 4.1 Proximal gradient method

Most work on learning with regularizers, including this work, can be thought of as instances of the *proximal gradient* method (Parikh and Boyd, 2014). Our objective function can be split into two parts, a convex and differentiable part ($L$) and a convex but non-differentiable part ($\lambda R$). In proximal gradient descent, we alternate between improving $L$ alone and $\lambda R$ alone. Let $\mathbf{u}$ be the parameter values from the previous iteration. We compute new parameter values $\mathbf{w}$ using:

$$\mathbf{v} \leftarrow \mathbf{u} - \eta \nabla L(\mathbf{u}) \quad (3)$$

$$\mathbf{w} \leftarrow \arg\max_{\mathbf{w}} \left(\frac{1}{2\eta}\|\mathbf{w} - \mathbf{v}\|^2 + \lambda R(\mathbf{w})\right) \quad (4)$$

and repeat until convergence. The first update is just a standard gradient descent update on $L$; the second is known as the *proximal operator* for $\lambda R$ and in many cases has a closed-form solution. In the rest of this section, we provide some justification for this method, and in Sections 4.2 and 4.3 we show how to compute the proximal operator for the $\ell_2$ and $\ell_\infty$ norms.

We can think of the gradient descent update (3) on $L$ as follows. Approximate $L$ around $\mathbf{u}$ by the tangent plane,

$$\bar{L}(\mathbf{v}) = L(\mathbf{u}) + \nabla L(\mathbf{u})(\mathbf{v} - \mathbf{u}) \quad (5)$$

and move $\mathbf{v}$ to minimize $\bar{L}$, but don't move it too far from $\mathbf{u}$; that is, minimize

$$F(\mathbf{v}) = \frac{1}{2\eta}\|\mathbf{v} - \mathbf{u}\|^2 + \bar{L}(\mathbf{v}).$$

Setting partial derivatives to zero, we get

$$\frac{\partial F}{\partial \mathbf{v}} = \frac{1}{\eta}(\mathbf{v} - \mathbf{u}) + \nabla L(\mathbf{u}) = 0$$

$$\mathbf{v} = \mathbf{u} - \eta \nabla L(\mathbf{u}).$$

By a similar strategy, we can derive the second step (4). Again we want to move $\mathbf{w}$ to minimize the objective function, but don't want to move it too far from $\mathbf{u}$; that is, we want to minimize:

$$G(\mathbf{w}) = \frac{1}{2\eta}\|\mathbf{w} - \mathbf{u}\|^2 + \bar{L}(\mathbf{w}) + \lambda R(\mathbf{w}).$$

Note that we have *not* approximated $R$ by a tangent plane. We can simplify this by substituting in (3). The first term becomes

$$\frac{1}{2\eta}\|\mathbf{w} - \mathbf{u}\|^2 = \frac{1}{2\eta}\|\mathbf{w} - \mathbf{v} - \eta\nabla L(\mathbf{u})\|^2$$
$$= \frac{1}{2\eta}\|\mathbf{w} - \mathbf{v}\|^2 - \nabla L(\mathbf{u})(\mathbf{w} - \mathbf{v})$$
$$+ \frac{\eta}{2}\|\nabla L(\mathbf{u})\|^2$$

and the second term becomes

$$\bar{L}(\mathbf{w}) = L(\mathbf{u}) + \nabla L(\mathbf{u})(\mathbf{w} - \mathbf{u})$$
$$= L(\mathbf{u}) + \nabla L(\mathbf{u})(\mathbf{w} - \mathbf{v} - \eta \nabla L(\mathbf{u})).$$

The $\nabla L(\mathbf{u})(\mathbf{w} - \mathbf{v})$ terms cancel out, and we can ignore terms not involving $\mathbf{w}$, giving

$$G(\mathbf{w}) = \frac{1}{2\eta}\|\mathbf{w} - \mathbf{v}\|^2 + \lambda R(\mathbf{w}) + \text{const.}$$

which is minimized by the update (4). Thus, we have split the optimization step into two easier steps: first, do the update for $L$ (3), then do the update for $\lambda R$ (4). The latter can often be done exactly (without approximating $R$ by a tangent plane). We show next how to do this for the $\ell_2$ and $\ell_\infty$ norms.

### 4.2 $\ell_2$ and $\ell_{2,1}$ regularization

Since the $\ell_{2,1}$ norm on matrices (1) is separable into the $\ell_2$ norm of each row, we can treat each row separately. Thus, for simplicity, assume that we have a single row and want to minimize

$$G(\mathbf{w}) = \frac{1}{2\eta}\|\mathbf{w} - \mathbf{v}\|^2 + \lambda\|\mathbf{w}\| + \text{const.}$$

The minimum is either at $\mathbf{w} = 0$ (the tip of the cone) or where the partial derivatives are zero (Figure 3):

$$\frac{\partial G}{\partial \mathbf{w}} = \frac{1}{\eta}(\mathbf{w} - \mathbf{v}) + \lambda \frac{\mathbf{w}}{\|\mathbf{w}\|} = 0.$$

Clearly, $\mathbf{w}$ and $\mathbf{v}$ must have the same direction and differ only in magnitude, that is, $\mathbf{w} = \alpha \frac{\mathbf{v}}{\|\mathbf{v}\|}$. Substituting this into the above equation, we get the solution

$$\alpha = \|\mathbf{v}\| - \eta\lambda.$$

Therefore the update is

$$\mathbf{w} = \alpha \frac{\mathbf{v}}{\|\mathbf{v}\|}$$
$$\alpha = \max(0, \|\mathbf{v}\| - \eta\lambda).$$

### 4.3 $\ell_\infty$ and $\ell_{\infty,1}$ regularization

As above, since the $\ell_{\infty,1}$ norm on matrices (2) is separable into the $\ell_\infty$ norm of each row, we can treat each row separately; thus, we want to minimize

$$G(\mathbf{w}) = \frac{1}{2\eta}\|\mathbf{w} - \mathbf{v}\|^2 + \lambda \max_j |x_j| + \text{const.}$$

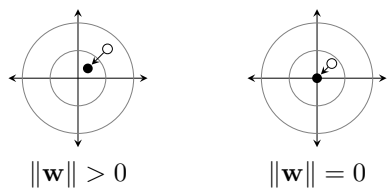

$\|\mathbf{w}\| > 0$ $\qquad$ $\|\mathbf{w}\| = 0$

Figure 3: Examples of the two possible cases for the $\ell_2$ gradient update. Point $\mathbf{v}$ is drawn with a hollow dot, and point $\mathbf{w}$ is drawn with a solid dot.

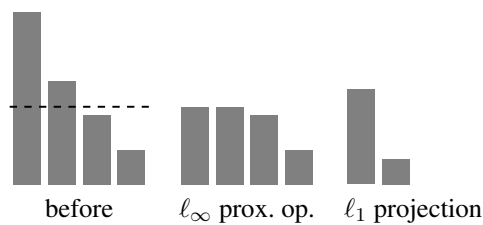

before $\qquad$ $\ell_\infty$ prox. op. $\qquad$ $\ell_1$ projection

Figure 4: The proximal operator for the $\ell_\infty$ norm (with strength $\eta\lambda$) decreases the maximal components until the total decrease sums to $\eta\lambda$. Projection onto the $\ell_1$-ball (of radius $\eta\lambda$) decreases each component by an equal amount until they sum to $\eta\lambda$.

Intuitively, the solution can be characterized as: Decrease all of the maximal $|x_j|$ until the total decrease reaches $\eta\lambda$ or all the $x_j$ are zero. See Figure 4.

If we pre-sort the $|x_j|$ in nonincreasing order, it's easy to see how to compute this: for $\rho = 1, \ldots, n$, see if there is a value $\xi \leq x_\rho$ such that decreasing all the $x_1, \ldots, x_\rho$ to $\xi$ amounts to a total decrease of $\eta\lambda$. The largest $\rho$ for which this is possible gives the correct solution.

But this situation seems similar to another optimization problem, projection onto the $\ell_1$-ball, which Duchi et al. (2008) solve in linear time without pre-sorting. In fact, the two problems can be solved by nearly identical algorithms, because they are convex conjugates of each other (Duchi and Singer, 2009; Bach et al., 2012). Intuitively, the $\ell_1$ projection of $v$ is exactly what is cut out by the $\ell_\infty$ proximal operator, and vice versa (Figure 4).

Duchi et al.'s algorithm modified for the present problem is shown as Algorithm 1. It partitions the $x_j$ about a pivot element (line 6) and tests whether it and the elements to its left can be decreased to a value $\xi$ such that the total decrease is $\delta$ (line 8). If so, it recursively searches the right side; if not, the left side. At the conclusion of the algorithm, $\rho$ is set to the largest value that passes the test (line 13), and finally the new $x_j$ are computed (line 16) – the only difference from Duchi et al.'s algorithm.

This algorithm is asymptotically faster than that of Quattoni et al. (2009). They reformulate $\ell_{\infty,1}$ regularization as a constrained optimization problem (in which the $\ell_{\infty,1}$ norm is bounded by $\mu$) and provide a solution in $\mathcal{O}(n \log n)$ time. The method shown here is simpler and faster because it can work on each row separately.

---

**Algorithm 1** Linear-time algorithm for the proximal operator of the $\ell_\infty$ norm.

1: **procedure** UPDATE($\mathbf{w}, \delta$)
2:    $lo, hi \leftarrow 1, n$
3:    $s \leftarrow 0$
4:    **while** $lo \leq hi$ **do**
5:       select $md$ randomly from $lo, \ldots, hi$
6:       $\rho \leftarrow$ PARTITION($\mathbf{w}, lo, md, hi$)
7:       $\xi \leftarrow \frac{1}{\rho}\left(s + \sum_{i=lo}^{\rho} |x_i| - \delta\right)$
8:       **if** $\xi \leq |x_\rho|$ **then**
9:          $s \leftarrow s + \sum_{i=lo}^{\rho} |x_i|$
10:         $lo \leftarrow \rho + 1$
11:      **else**
12:         $hi \leftarrow \rho - 1$
13:    $\rho \leftarrow hi$
14:    $\xi \leftarrow \frac{1}{\rho}(s - \delta)$
15:    **for** $i \leftarrow 1, \ldots, n$ **do**
16:       $x_i \leftarrow \min(\max(x_i, -\xi), \xi)$
17: **procedure** PARTITION($\mathbf{w}, lo, md, hi$)
18:    swap $x_{lo}$ and $x_{md}$
19:    $i \leftarrow lo + 1$
20:    **for** $j \leftarrow lo + 1, \ldots, hi$ **do**
21:       **if** $x_j \geq x_{lo}$ **then**
22:          swap $x_i$ and $x_j$
23:          $i \leftarrow i + 1$
24:    swap $x_{lo}$ and $x_{i-1}$
25:    **return** $i - 1$

---

## 5 Experiments

We evaluate our model using the open-source NPLM toolkit released by Vaswani et al. (2013), extending it to use the additional regularizers as described in this paper.[2] We use a vocabulary size of 100k and word embeddings with 50 dimensions. We use two hidden layers of rectified linear units (Nair and Hinton, 2010).

---
[2]These extensions have been contributed to the NPLM project.

We train neural language models (LMs) on two natural language corpora, Europarl v7 English and the AFP portion of English Gigaword 5. After tokenization, Europarl has 56M tokens and Gigaword AFP has 870M tokens. For both corpora, we hold out a validation set of 5,000 tokens. We train each model for 10 iterations over the training data.

Our experiments break down into three parts. First, we look at the impact of our pruning method on perplexity of a held-out validation set, across a variety of settings. Second, we take a closer look at how the model evolves through the training process. Finally, we explore the downstream impact of our method on a statistical phrase-based machine translation system.

### 5.1 Evaluating perplexity and network size

We first look at the impact that the $\ell_{\infty,1}$ regularizer has on the perplexity of our validation set. The main results are shown in Table 1. For $\lambda \leq 0.01$, the regularizer seems to have little impact: no hidden units are pruned, and perplexity is also not affected. For $\lambda = 1$, on the other hand, most hidden units are pruned – apparently too many, since perplexity is worse. But for $\lambda = 0.1$, we see that we are able to prune out many hidden units: up to half of the first layer, with little impact on perplexity. We found this to be consistent across all our experiments, varying $n$-gram size, initial hidden layer size, and vocabulary size.

Table 2 shows the same information for 5-gram models trained on the larger Gigaword AFP corpus. These numbers look very similar to those on Europarl: again $\lambda = 0.1$ works best, and, counter to expectation, even the final number of units is similar.

Table 3 shows the result of varying the vocabulary size: again $\lambda = 0.1$ works best, and, although it is not shown in the table, we also found that the final number of units did not depend strongly on the vocabulary size.

Table 4 shows results using the $\ell_{2,1}$ norm (Europarl corpus, 5-grams, 100k vocabulary). Since this is a different regularizer, there isn't any reason to expect that $\lambda$ behaves the same way, and indeed, a smaller value of $\lambda$ seems to work best.

### 5.2 A closer look at training

We also studied the evolution of the network over the training process to gain some insights into how the method works. The first question we want to answer is whether the method is simply removing units, or converging on an optimal number of units. Figure 5 suggests that it is a little of both: if we start with too many units (900 or 1000), the method converges to the same number regardless of how many extra units there were initially. But if we start with a smaller number of units, the method still prunes away about 50 units.

|       | 2-gram |         |     | 3-gram |         |     | 5-gram |         |     |
|-------|--------|---------|-----|--------|---------|-----|--------|---------|-----|
| λ     | layer 1 | layer 2 | ppl | layer 1 | layer 2 | ppl | layer 1 | layer 2 | ppl |
| 0     | 1,000  | 50      | 103 | 1,000  | 50      | 66  | 1,000  | 50      | 55  |
| 0.001 | 1,000  | 50      | 104 | 1,000  | 50      | 66  | 1,000  | 50      | 54  |
| 0.01  | 1,000  | 50      | 104 | 1,000  | 50      | 63  | 1,000  | 50      | 55  |
| 0.1   | 499    | 47      | 105 | 652    | 49      | 66  | 784    | 50      | 55  |
| 1.0   | 50     | 24      | 111 | 128    | 32      | 76  | 144    | 29      | 68  |

Table 1: Comparison of $\ell_{\infty,1}$ regularization on 2-gram, 3-gram, and 5-gram neural language models. The network initially started with 1,000 units in the first hidden layer and 50 in the second. A regularization strength of $\lambda = 0.1$ consistently is able to prune units while maintaining perplexity, even though the final number of units varies considerably across models. The vocabulary size is 100k.

| λ     | layer 1 | layer 2 | perplexity |
|-------|---------|---------|------------|
| 0     | 1,000   | 50      | 100        |
| 0.001 | 1,000   | 50      | 99         |
| 0.01  | 1,000   | 50      | 101        |
| 0.1   | 742     | 50      | 107        |
| 1.0   | 24      | 17      | 173        |

Table 2: Results from training a 5-gram neural LM on the AFP portion of the Gigaword dataset. As with the smaller Europarl corpus (Table 1), a regularization strength of $\lambda = 0.1$ is able to prune units while maintaining perplexity.

|       | vocabulary size |     |     |      |
|-------|-----------------|-----|-----|------|
| λ     | 10k             | 25k | 50k | 100k |
| 0     | 47              | 60  | 54  | 55   |
| 0.001 | 47              | 54  | 54  | 54   |
| 0.01  | 47              | 58  | 55  | 55   |
| 0.1   | 48              | 62  | 55  | 55   |
| 1.0   | 61              | 64  | 65  | 68   |

Table 3: A regularization strength of $\lambda = 0.1$ is best across different vocabulary sizes.

| λ      | layer 1 | layer 2 | perplexity |
|--------|---------|---------|------------|
| 0      | 1,000   | 50      | 100        |
| 0.0001 | 1,000   | 50      | 54         |
| 0.001  | 1,000   | 50      | 55         |
| 0.01   | 616     | 50      | 57         |
| 0.1    | 199     | 32      | 65         |

Table 4: Results using $\ell_{2,1}$ regularization.

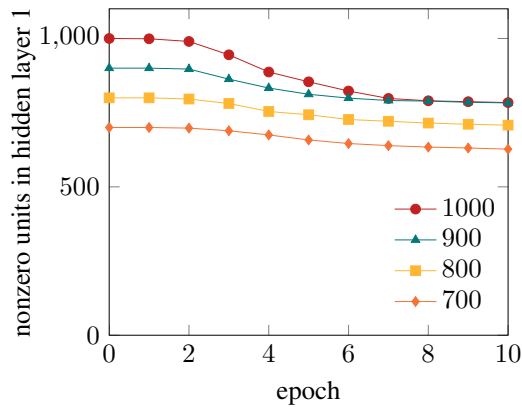

Figure 5: Number of units in first hidden layer over time, with various starting sizes ($\lambda = 0.1$). If we start with too many units, we end up with the same number, although if we start with a smaller number of units, a few are still pruned away.

Next, we look at the behavior over time of different regularization strengths $\lambda$. We found that not only does $\lambda = 1$ prune out too many units, it does so at the very first iteration (Figure 6, above), perhaps prematurely. By contrast, the $\lambda = 0.1$ run prunes out units gradually. By plotting these curves together with perplexity (Figure 6, below), we can see that the $\lambda = 0.1$ run is fitting the model and pruning it at the same time, which seems preferable to fitting without any pruning ($\lambda =$

| $\lambda$ | none | neural LM Europarl | Gigaword AFP |
|---|---|---|---|
| 0 (none) | 23.2 | 24.7 (+1.5) | 25.2 (+2.0) |
| 0.1 | | 24.6 (+1.4) | 24.9 (+1.7) |

Table 5: The improvements in translation accuracy due to the neural LM (shown in parentheses) are affected only slightly by $\ell_{\infty,1}$ regularization. For the Europarl LM, there is no statistically significant difference, and for the Gigaword AFP LM, a statistically significant but small decrease of $-0.3$.

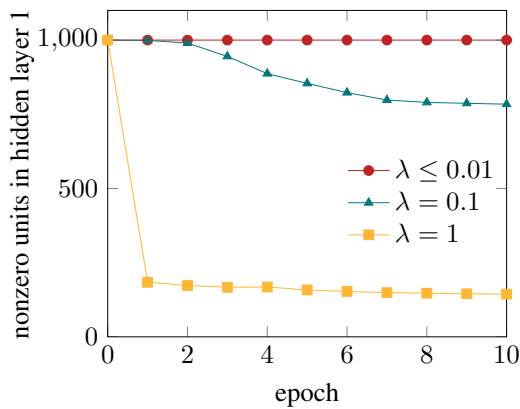

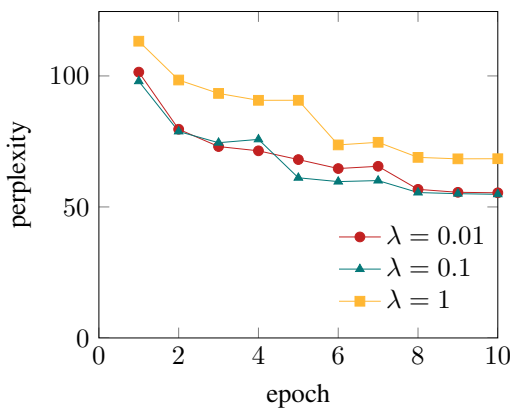

Figure 6: Above: Number of units in first hidden layer over time, for various regularization strengths $\lambda$. A regularization strength of $\leq 0.01$ does not zero out any rows, while a strength of 1 zeros out rows right away. Below: Perplexity over time. The runs with $\lambda \leq 0.1$ have very similar learning curves, whereas $\lambda = 1$ is worse from the beginning.

0.01) or pruning first and then fitting ($\lambda = 1$).

We can also visualize the weight matrix itself over time (Figure 7), for $\lambda = 0.1$. It is striking that although this setting fits the model and prunes it at the same time, as argued above, by the first iteration it already seems to have decided roughly how many units it will eventually prune.

### 5.3 Evaluating on machine translation

We also looked at the impact of our method on statistical machine translation systems. We used the Moses toolkit (Koehn et al., 2007) to build a phrase based machine translation system with a traditional 5-gram LM trained on the target side of our bitext. We augmented this system with neural LMs trained on the Europarl data and the Gigaword AFP data. Based on the results from the perplexity experiments, we looked at models both built with a $\lambda = 0.1$ regularizer, and without regularization ($\lambda = 0$).

We built our system using the newscommentary dataset v8. We tuned our model using newstest13 and evaluated using newstest14. After standard cleaning and tokenization, there were 155k parallel sentences in the newscommentary dataset, and 3,000 sentences each for the tuning and test sets.

Table 5 shows that the addition of a neural LM helps substantially over the baseline, with improvements of up to 2 BLEU. Using the Europarl model, the BLEU scores obtained without and with regularization were not significantly different ($p \geq 0.05$), consistent with the negligible perplexity difference between these models. On the Gigaword AFP model, regularization did decrease the BLEU score by $0.3$, consistent with the small perplexity increase of the regularized model. The decrease is statistically significant, but small compared with the overall benefit of adding a neural LM.

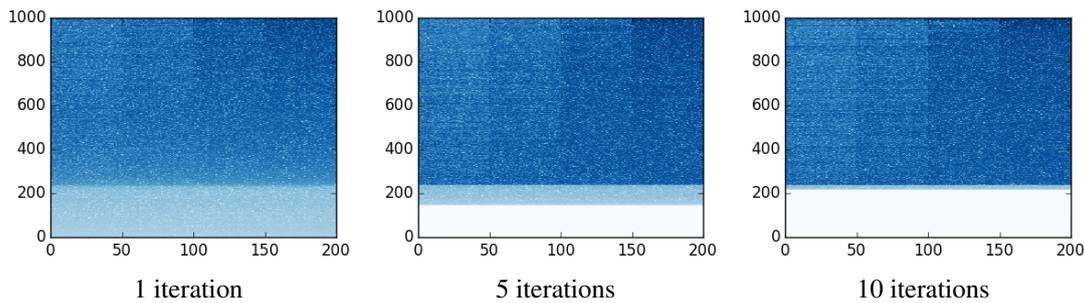

Figure 7: Evolution of the first hidden layer weight matrix after 1, 5, and 10 iterations (with rows sorted by $\ell_\infty$ norm). A nonlinear color scale is used to show small values more clearly. The four vertical blocks correspond to the four context words. The light bar at the bottom is the rows that are close to zero, and the white bar is the rows that are exactly zero.

## 6 Related Work

Researchers have been exploring the use of neural networks for language modeling for a long time. Schmidhuber and Heil (1996) proposed a character $n$-gram model using neural networks which they used for text compression. Xu and Rudnicky (2000) proposed a word-based probability model using a softmax output layer trained using cross-entropy, but only for bigrams. Bengio et al. (2003) defined a probabilistic word $n$-gram model and demonstrated improvements over conventional smoothed language models. Mnih and Teh (2012) sped up training of log-bilinear language models through the use of noise-contrastive estimation (NCE). Vaswani et al. (2013) also used NCE to train the architecture of Bengio et al. (2003), and were able to integrate a large-vocabulary language model directly into a machine translation decoder. Baltescu et al. (2014) describe a similar model, with extensions like a hierarchical softmax (based on Brown clustering) and direct $n$-gram features.

Beyond feed-forward neural network language models, researchers have explored using more complicated neural network architectures. RNNLM is an open-source implementation of a language model using recurrent neural networks (RNN) where connections between units can form directed cycles (Mikolov et al., 2011). Sundermeyer et al. (2015) use the long-short term memory (LSTM) neural architecture to show a perplexity improvement over the RNNLM toolkit. In future work, we plan on exploring how our method could improve these more complicated neural models as well.

Automatically limiting the size of neural networks is an old idea. The "Optimal Brain Damage" (OBD) technique (LeCun et al., 1989) computes a *saliency* based on the second derivative of the objective function with respect to each parameter. The parameters are then sorted by saliency, and the lowest-saliency parameters are pruned. The pruning process is separate from the training process, whereas regularization performs training and pruning simultaneously. Regularization in neural networks is also an old idea; for example, Nowland and Hinton (1992) mention both $\ell_2^2$ and $\ell_0$ regularization. Our method develops on this idea by using a mixed norm to prune units, rather than parameters.

Srivastava et al. introduce a method called *dropout* in which units are directly deactivated at random during training (Srivastava et al., 2014), which induces sparsity in the hidden unit activations. However, at the end of training, all units are reactivated, as the goal of dropout is to reduce overfitting, not to reduce network size. Thus, dropout and our method seem to be complementary.

## 7 Conclusion

We have presented a method for auto-sizing a neural network during training by removing units using a $\ell_{\infty,1}$ regularizer. This regularizer drives a unit's input weights as a group down to zero, allowing the unit to be pruned. We can thus prune units out of our network during training with minimal impact to held-out perplexity or downstream performance of a machine translation system.

Our results showed empirically that the choice of a regularization coefficient of 0.1 was robust to initial configuration parameters of initial network size, vocabulary size, $n$-gram order, and training corpus. Furthermore, imposing a single regularizer on the objective function can tune all of the hidden layers of a network with one setting. This reduces the need to conduct expensive, multi-dimensional grid searches in order to determine optimal sizes.

We have demonstrated the power and efficacy of this method on a feed-forward neural network for language modeling though experiments on perplexity and machine translation. However, this method is general enough that it should be applicable to other domains, both inside natural language processing and outside. As neural models become more pervasive in natural language processing, the ability to auto-size networks for fast experimentation and quick exploration will become increasingly important.

## Acknowledgments

We would like to thank Tomer Levinboim, Antonios Anastasopoulos, and Ashish Vaswani for their helpful discussions, as well as the reviewers for their assistance and feedback.